\documentclass{article}

    \PassOptionsToPackage{numbers, compress}{natbib}

\usepackage[final]{neurips_2024}

\usepackage[utf8]{inputenc} 
\usepackage[T1]{fontenc}    
\usepackage{hyperref}       
\usepackage{url}            
\usepackage{booktabs}       
\usepackage{amsfonts}       
\usepackage{nicefrac}       
\usepackage{microtype}      
\usepackage{xcolor}         
\usepackage{graphicx} 
\usepackage{caption}
\usepackage{amsmath} 
\usepackage{array}
\title{Towards LLM-guided Efficient and Interpretable Multi-linear Tensor Network Rank Selection}

\author{
  Giorgos Iacovides\thanks{Equal contribution.},\; Wuyang Zhou\footnotemark[1],\; Danilo Mandic \\
  Department of Electrical and Electronic Engineering \\
  Imperial College London \\
  \texttt{\{giorgos.iacovides20, wuyang.zhou19, d.mandic\}@imperial.ac.uk} \\
}

\begin{document}

\maketitle

\begin{abstract}
We propose a novel framework that leverages large language models (LLMs) to guide the rank selection in tensor network models for higher-order data analysis. By utilising the intrinsic reasoning capabilities and domain knowledge of LLMs, our approach offers enhanced interpretability of the rank choices and can effectively optimise the objective function. This framework enables users without specialised domain expertise to utilise tensor network decompositions and understand the underlying rationale within the rank selection process. Experimental results validate our method on financial higher-order datasets, demonstrating interpretable reasoning, strong generalisation to unseen test data, and its potential for self-enhancement over successive iterations. This work is placed at the intersection of large language models and higher-order data analysis.
\end{abstract}

\section{Introduction}
The exponential increase in the volume and richness of available data has led to a widespread use of higher-order data, often represented as higher-order tensors. Tensor decompositions aim to break down these high-dimensional tensors into simpler components, effectively capturing their latent patterns and correlations. These techniques have been applied across various fields, such as machine learning, signal processing, computer vision, and quantum physics \cite{td_1,td_2,td_3,td_4,td_5,td_6,td_7,td_8}. The success of tensor decomposition techniques is closely linked to their ability to mitigate the curse of dimensionality. By carefully designing the structure of core tensors and multi-linear operations among them, many tensor decomposition algorithms have emerged. Of particular interest is the Fully Connected Tensor Network (FCTN) decomposition, which decomposes an  $N$\textit{th}-order tensor into $N$ small-sized $N$\textit{th}-order core tensors and captures the correlation between any two tensor modes \cite{fctn}. \par 
However, practitioners of tensor decompositions face significant challenges related to model selection, particularly in determining the optimal tensor ranks — a problem known as rank selection (RS). Specifically, finding these optimal tensor ranks has been proven to be NP-hard \cite{np_hard, tnale}, and for most practical problems, brute-force rank search is infeasible due to the `combinatorial explosion' \cite{tnale}, which results in prohibitively high computational costs and time requirements. \par 
As a result, in most studies \cite{td_app_1, td_app_2,td_app_3,td_app_5}, the rank is either treated as a fixed hyperparameter, set by the author based on domain knowledge, or determined through random search over possible values. Although recent works have attempted to determine the optimal rank using statistical or sampling-based optimisation methods, these typically address other tensor decomposition problems \cite{tnls, adap_rank_1, adap_rank_2}. In contrast, the FCTN decomposition poses a more challenging problem for rank selection due to its fully connected structure, thus facing an exponential growth in the number of possible rank combinations when the number of modes increases. \par
At the same time, the rise of transformer-based LLMs pretrained on vast text corpora has demonstrated a remarkable capacity for `reasoning' \cite{large_llm}. This reasoning ability is further enhanced when LLMs are guided by task-specific prompting strategies \cite{reason_1,reason_2,reason_3,reason_4}, and they have shown exceptional performance in complex question-answering and prediction tasks that require real-world knowledge \cite{qa_1,qa_2}. These findings suggest that, through pretraining on diverse textual data, LLMs encode rich knowledge about real-world relationships, which they can effectively leverage to perform various downstream tasks \cite{encoder_1}. It is noteworthy that as recently as Septemeber 2024, Open AI released o1, a new LLM trained with reinforcement learning to perform complex reasoning \cite{openai_new}. This highlights the growing interest and focus on reasoning with LLMs, which promises to advance the task of automatic evaluation with these models.
\par 

To the best of our knowledge, only a few works have specifically addressed the rank selection problem for the FCTN \cite{tnale}\cite{svdins_tn}\cite{latent_factor}\cite{stochastic_svd}. However, these methods all fall short in one key aspect: they lack interpretability regarding the selection of ranks. Thus, the fundamental question we address in this paper is:
\begin{itemize}
\item  Can we establish enhanced interpretability in the tensor network rank selection process using the reasoning ability and domain knowledge of LLMs?
\end{itemize}
To this end, we propose a framework for LLM-guided tensor network rank selection that enhances the interpretability of the rank selection process. By leveraging the ability of LLMs to analyse complex interactions between tensor modes and incorporating their inherent domain knowledge, our approach  provides a unique, transparent, interpretable and efficient rank selection method. This framework also enables users without domain-specific expertise to easily determine tensor network ranks and understand the underlying reasoning behind their selection. Experimental results show that our proposed framework outperforms the baseline models.
\par 
The main contributions of this paper are three-fold: 
\begin{itemize}
    \item We design an LLM-guided tensor network rank selection framework which achieves high interpretability and good generalisation abilities based on the inherent domain knowledge of the LLM;
    \item The analysis of the responses of the LLM enables a better understanding of the connections between the ranks and the intrinsic higher-order mode interactions in a tensor;
    \item To the best of our knowledge, this work is the first to propose a systematic framework which utilises LLMs to estimate the tensor network ranks directly based on iterative reasoning.
\end{itemize}

\section{Tensor Network Rank Search and Its Search Space}
\begin{figure}[!t]
    \centering
    \includegraphics[width=0.6\hsize]{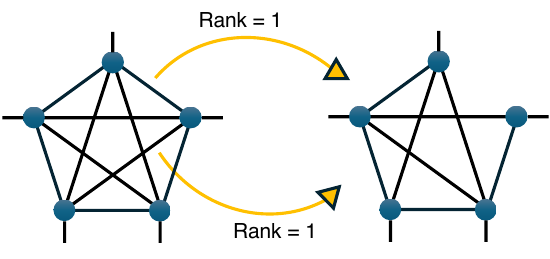}
    \caption{Setting a rank in the FCTN topology to $1$ is equivalent to dropping that connection, resulting in a new tensor network structure. In this way, tensor network rank search for FCTN becomes equivalent to tensor network structure search under the constraint of a complete graph.}
    \label{fig:rs2ss}
\end{figure}

Tensor networks can be represented using the graphical notation \cite{orus2014practical}, whereby a tensor network is represented using a set of vertices and edges. Each vertex represents a decomposed core tensor, and the closed edges between two core tensors are generalised higher-order matrix multiplications, termed tensor contractions \cite{fctn}. Each closed edge has an assigned value, its rank, which indicates the degree of connection between the two vertices. By setting the rank of a closed edge to $1$, we can harmlessly drop that edge, resulting in a new tensor network structure. Thus, as shown in Figure \ref{fig:rs2ss}, the tensor network structure search problem is equivalent to tensor network rank search under a given fixed number of vertices \cite{li2021rank}.

Since the FCTN \footnote{A detailed mathematical definition of the FCTN decomposition can be found in section \ref{sec:fctn}.} topology can exploit mode interactions between any two tensor modes, we constrain the search space to a fully connected tensor network \cite{fctn}, which has a graphical notation resembling a complete graph $G = (V, E)$ \cite{li2021rank, ye2018tensor, harary2018graph}, where $G$ denotes the graph, $V$ denotes the $N$ core tensor vertices of the $N$th-order tensor, $\mathcal{X} \in \mathbb{R}^{I_1 \times I_2 \times \ldots \times I_N}$, and $E$ denotes the rank values of each edge. In other words, the tensor network structure is constrained to having $N$ vertices for an $N$-order tensor, and every vertex is fully connected to all other vertices. In this way, the search for the ranks of the FCTN topology is equivalent to TN structure search under the constraint of a complete graph \cite{li2021rank}, which generalises the rank search and mode permutation search \cite{tnls} of popular TN decompositions, such as the Tensor Train \cite{oseledets2011tensor} and the Tensor Ring \cite{zhao2016tensor} decompositions. 

\paragraph{Objective Function.} In this work, we focus specifically on the tensor compression task, which aims to balance between approximating the higher-order data tensors using low-dimensional features as well as possible, while compressing the data as efficiently as possible. Increasing the ranks usually leads to decreases in the approximation errors, but it also leads to a worse compression ratio (CR), defined as the total number of parameters in the core tensors divided by the total number of parameters in the original tensor. We formalise the objective function to be a minimisation of the loss function, which is a combination of the complexity of the tensor network structure and the approximation error, similar to \cite{tnale, zengtngps}, and is given by

\begin{equation} \label{loss}
    \min_E \log_{10}(\text{CR}) + \frac{ \lambda \sum_{n=1}^{N} \left( \frac{\| \mathcal{X}_n - \mathcal{X}_{n, approx} \|_F}{\| \mathcal{X}_n \|_F} \right)}{N} \ \ s.t. \ \ \mathcal{X}_{n, approx} \in G \ \ \text{for} \ \ n = 1, 2, \ldots, N.
\end{equation}

where $\text{CR}= \frac{\text{Total number of parameters in the core tensors}}{\text{Total number of parameters in the original tensor}} $, $\lambda$ is a scaling factor set as $10^3$, $N$ is the total number of tensors to be compressed, $\mathcal{X}_n$ is the $n$-th original tensor, and $\mathcal{X}_{n, approx}$ is the approximated $n$-th tensor obtained from the FCTN core tensors using the FCTN-ALS algorithm \cite{fctn}.

\section{LLM-guided Interpretable Rank Selection} \label{llm_info}
For our prompting strategy, the popular `chat' interface is employed, where interactions with the LLM follow a structured dialogue between the \textit{user} and the \textit{assistant} \cite{chat_1,chat_2}. The user message serves as the prompt that we supply to the LLM, while the assistant message represents the response of the model. This dialogue-based approach is intuitive for generating stepwise, conversational reasoning, making it particularly useful for iterative tasks such as rank selection in tensor decomposition. \par 
Moreover, in our approach we prompt the model with the entire conversational history of messages up to that point. This enables the LLM to retain context from previous steps, allowing it to better understand the problem and refine its responses in subsequent iterations.  We postulate that this accumulation of context provides significant benefits by maintaining continuity, ensuring the relevance of the reasoning of the model, and enabling more informed and accurate responses as the experiment progresses. \par 
It is important to note that this approach comes at a cost, as it leads to a linear increase in the number of messages with each iteration, as every iteration introduces two additional messages: the user prompt and the response of the LLM. Consequently, it is crucial to ensure that the total length of the accumulated input, which includes all user prompts and LLM responses across iterations, and output, remains within the context window limit allowed by the model. \par 
We now describe the specific details of our system message, input prompt, and iterative prompt that enable the LLM to make informed decisions regarding rank selection. Real-world data experiments show that these components leads to improved effectiveness in the rank selections, in terms of minimising the loss function and enhancing the interpretability of the responses of LLMs. \par 
\paragraph{System Message.}
The `system message' refers to the prompt for setting the context or guiding the behavior of the LLM during a conversation. It is the first message in a dialogue and is used to establish the role which the LLM should play in the conversation. The following guidelines were applied to structure the system message: 
\begin{enumerate}
    \item Set the context by framing the LLM as a domain expert in the specific area of application;
    \item Encourage the model to use domain-knowledge and define explicitly the expected output;
    \item Define a clear optimisation objective that the model needs to solve. e.g. `minimise the loss function defined in Equation \ref{loss}';
    \item  Define how the different components of the loss function are calculated and explain how they impact the rank selection decisions;
    \item  Encourage the model to \textbf{explore} rank reduction if accuracy permits, to further optimise compression; this prevents the model from getting stuck in local minima where the approximation error is already negligible, allowing it to avoid overestimating ranks and enabling further compression without sacrificing accuracy;
    \item  Ask the model to provide step-by-step reasoning about its decisions.
\end{enumerate}

\par

\begin{figure}[!t]
    \centering
    \includegraphics[width=1\hsize]{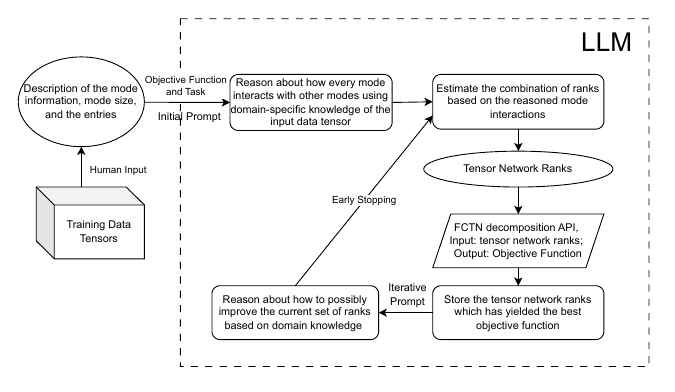}
    \caption{The proposed framework for LLM-guided tensor network rank search. The LLM first uses domain knowledge to reason about the mode interactions and then suggests the tensor network ranks. After obtaining the objective function values, the LLM tries to revise its reasoning and proposes a new set of tensor network ranks to optimise the objective function in an iterative cycle.}
    \label{fig:framework}
\end{figure}

\paragraph{Initial Prompt.} The initial prompt is the first message provided by the user to the LLM and is designed to frame the problem and guide the thought process of the model. The following guidelines were applied to structure the initial prompt:
\begin{enumerate}
    \item  Describe the structure of the tensor clearly, including the number of modes and the specific domain data dimensions involved;
    \item Define the task of rank selection and the objective to minimise the loss function stated in Equation \ref{loss};
    \item Instruct the model to reason step-by-step and utilise domain knowledge to reason the strength of interactions between modes;
    \item Explain to the model how to relate the intrinsic interactions between modes to the rank selection process;
    \item Specify the exact format in which the model should output its reasoning and suggested ranks;
    \item Impose practical constraints to ensure the ranks selected are meaningful.
\end{enumerate}

\paragraph{Iterative Prompt.} 
The iterative prompt is the message supplied by the user to the LLM in all iterations following the first. It guides the LLM through the rank refinement process by providing context from the previous and best iterations and encouraging exploration of new rank configurations to further minimise the loss function. The following guidelines were applied to structure the iterative prompt:
\begin{enumerate}
    \item Supply the model with both the previous and the best objective function values and the corresponding tensor network ranks to maintain context continuity and guide the rank adjustments;
    \item Remind the model of the specific goal to minimise the loss function specified in Equation \ref{loss};
    \item  Explain the relationship between rank changes and their impact on the different components of the loss function specified in Equation \ref{loss};
    \item Encourage the model to explore new rank configurations and prevent it from repeating previous ones;
    \item Guide the model to reason how different mode interactions could affect the overall decomposition accuracy;
    \item Specify the exact format in which the model should output its reasoning and suggested ranks;
    \item Impose practical constraints to ensure the ranks selected are meaningful.
\end{enumerate}

\section{Methodology}

As illustrated in Figure \ref{fig:framework}, our LLM-guided tensor network rank search framework operates in an iterative fashion. Initially, the LLM processes mode information about the input tensors, along with a system message and prompt, to recommend an initial set of tensor network ranks. These ranks are then passed through the FCTN decomposition API, which evaluates their effectiveness via the objective function. The LLM subsequently receives the iterative prompt, refines its reasoning, and generates new rank suggestions in a repeated cycle to optimise the objective function, until an early stopping condition has been met or it has reached 10 iterations.

\paragraph{Early Stopping.} Since there is no convergence guarantee in the rank values the LLM suggests, we employ an early stopping technique to stop the iterative cycle early if the objective function calculated on the training data has not improved for a certain number of iterations.

\paragraph{Rank Constraints.} The authors in \cite{svdins_tn} showed that the ranks in a FCTN decomposition are upper bounded by the sizes of the two modes, of which each edge connects. Therefore, an automatic check is implemented to verify that the LLM-suggested ranks do not exceed their upper bounds and if so, reduce them to the upper bound. This technique effectively reduces the rank search space for the LLM. This restriction is also applied to the range of the random search and Bayesian optimisation-based baselines detailed in Section \ref{sec:exp}.

\paragraph{Interpretability.} A major aim of our proposed framework is to enhance the interpretability in the rank selection process. In particular, by analyzing the reasoning in the LLM responses, we can understand why certain interactions require a specific rank. Most importantly, this allows any non-expert to decide on the tensor ranks effectively using LLMs. Furthermore, in cases where an expert has trouble in figuring out the mode interactions, LLMs can help guide the process.

\begin{figure}[!t]
\centering
\includegraphics[width=1\textwidth]{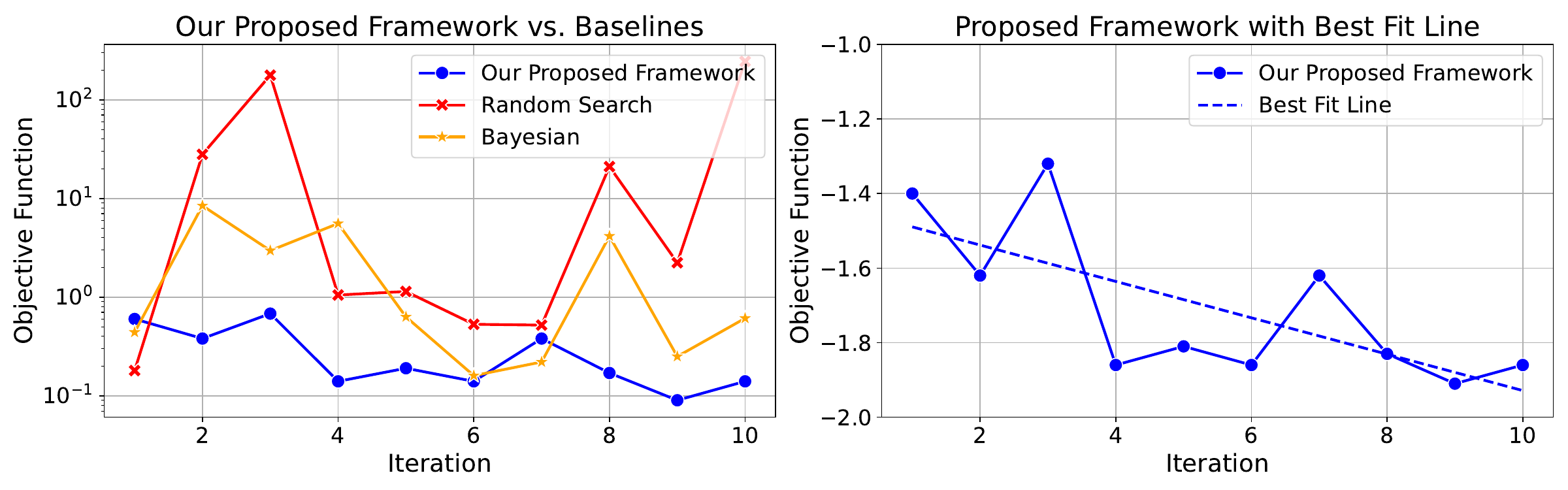}%
\label{fig:proposed_vs_random}
\caption{Visualisation of the experimental results; Left: Comparison of the objective function between our proposed framework and the baseline models across 10 iterations. Right: Evolution of the objective function for our proposed framework across 10 iterations with the line of best fit.}
\label{fig:results}
\end{figure}

\section{Experimental Results}
\label{sec:exp}

Our method can be flexibly applied to any tensor data across various domains. In the experiments, financial time series data were used, as they contain multidimensional complex interactions between different factors. These interactions make tensor decomposition particularly useful for revealing hidden patterns and correlations. Additionally, financial data is rich in cross-modal information \cite{scalzo2021analysis}, making it a great candidate for testing the interpretability and effectiveness of our proposed framework.

\paragraph{Data Preparation.} We structured financial time series data into 142 temporally ordered fifth-order tensors, denoted as $\{\mathcal{X}_n\}_{n=1}^{142} \in \mathbb{R}^{3 \times 6 \times 3 \times 4 \times 5}$, each representing a rolling window produced via multi-way delay embedding through the temporal direction \cite{yokota2018missing}. This leads to the value selection process of $10$ ranks. The first $80\%$ of these tensors were used as the training data, while the remaining $20\%$ with non-overlapping entries with the training data were used for testing. The modes of each tensor correspond to:

\begin{itemize} \item \textbf{Mode 1}: Types of financial instruments \item \textbf{Mode 2}: Asset indices within each type \item \textbf{Mode 3}: Features of each asset \item \textbf{Mode 4}: Interval lengths for averaging feature values \item \textbf{Mode 5}: Time points within each rolling window \end{itemize}

\paragraph{Experimental Settings.} The maximum number of iterations was set to 10, with an early stopping patience of 5 and a delta of zero. This means that if the objective function does not improve relative to the best one for five consecutive iterations, the process is stopped early. For the experiment, the LLM model \texttt{gpt-4o} \cite{HelloGPT4o} from OpenAI was chosen, as it supports a context window of up to 128,000 tokens and has broad domain-specific knowledge. To ensure that the LLM can provide comprehensive reasoning for all rank selections while avoiding context window overflow across iterations, as explained in Section \ref{llm_info}, the maximum output length was set to 3,000 tokens. 

\paragraph{Quantitative Results.} The quantitative results of our proposed framework are presented in Table \ref{tab:results} alongside the results from a random rank search algorithm and a Bayesian optimisation-based algorithm. Our framework outperformed the baseline models, achieving a lower overall objective function. Moreover, unlike the baseline models, in which the approximation error spikes at certain iterations, our LLM-guided rank search framework consistently maintains lower training and test approximation errors. This is because the LLM can make informed decisions based on its domain knowledge at every iteration, including the very first. We also note that the LLM-suggested rank values consistently yielded test approximation errors lower than the training errors after iteration 2, which is not observed in the baseline algorithms (see Appendix \ref{full_errors}). This demonstrates the generalisation ability of the proposed framework, as a result of the LLM's understanding of the interactions between modes in the training data, which effectively translates to the test data. Additionally, our proposed framework is able to provide the reasoning behind the rank selection choices, unlike the black-box nature of Bayesian optimisation. From the right panel of Figure \ref{fig:results}, observe that the overall trend of the objective function in our proposed framework decreases through successive iterations. This supports our hypothesis that the LLM applies its domain knowledge to continually refine its understanding of the intrinsic interactions between modes, leveraging insights from past iterations to adjust its rank selections accordingly. 
\begin{table}[ht]
\centering
\renewcommand{\arraystretch}{1.2}
\setlength{\tabcolsep}{3pt}
\begin{tabular}{|>{\centering\arraybackslash}p{2.3cm}|>{\centering\arraybackslash}p{0.9cm}|>{\centering\arraybackslash}p{0.9cm}|>{\centering\arraybackslash}p{0.9cm}|>{\centering\arraybackslash}p{0.9cm}|>{\centering\arraybackslash}p{0.9cm}|>{\centering\arraybackslash}p{0.9cm}|>{\centering\arraybackslash}p{0.9cm}|>{\centering\arraybackslash}p{0.9cm}|>{\centering\arraybackslash}p{0.9cm}|>{\centering\arraybackslash}p{0.9cm}|}

\hline
\textbf{Iteration} & 1 & 2 & 3 & 4 & 5 & 6 & 7 & 8 & 9 & 10 \\
\hline
\multicolumn{11}{|c|}{\textbf{Random Se}\textbf{arch}} \\
\hline
Train Obj. Func. & 25.98 & 175.44 & -0.95 & -1.82 &-0.86 & -1.47 & -1.48 & 19.04 & 0.23 & 243.06 \\
Train Error & --- & --- & --- & --- & --- & 0.2 & --- & --- & --- & --- \\
Test Obj. Func. & 29.7  & 217.2 & -0.88 & -1.82 &  -0.71 & -1.47 & -1.34  & 22.09 & 0.64 & 340.11 \\
Test Error & --- & --- & --- & --- & --- & 0.3 & --- & --- & --- & --- \\
Log CR &  -2.51 & -2.79 & -1.93 & -1.82 & -1.96 & -1.47 & -1.84 & -2.51 & -2.02 & -2.89 \\
\hline
\multicolumn{11}{|c|}{\textbf{Bayesian Optimisation}} \\
\hline
Train Obj. Func. & -1.56 & 6.45  & 1.00  & 3.57 & -1.37  & -1.84  & -1.78 & 2.15 & -1.75 & -1.39  \\
Train Error & --- & --- & --- & --- & 0.7 & 1.9 & 28.8 & --- & --- & 0.1 \\
Test Obj. Func. & -1.53 & 7.16 & 1.29 & 3.82 & -1.37 & -1.84 & -1.78  & 2.17 & -1.75& -1.39  \\
Test Error & ---  &---  & --- & --- & 1.0 & 1.7 & 15.6 & --- & --- & 0.1 \\
Log CR & -1.90 & -2.20 & -2.09 & -2.22 & -1.37 & -1.84 & -1.78 & -2.09  & -1.75  & -1.39\\
\hline
\multicolumn{11}{|c|}{\textbf{Our Proposed Framework}} \\
\hline
Train Obj. Func. & -1.40 & -1.62 & -1.32 & \underline{-1.86} & -1.81 & \underline{-1.86} & -1.62 & -1.83 & \textbf{-1.91} & \underline{-1.86} \\
Train Error & 0.4 & 1.8 & 0.8 & 11.9 & 9 & 10.7 & 1.8 & 7.8 & 22.9 & 11.2 \\
Test Obj. Func. & -1.40 & -1.62 & -1.32 & \underline{-1.86} & -1.81 & \underline{-1.86} & -1.62 & -1.83 & \textbf{-1.91} & \underline{-1.86} \\
Test Error & 0.5 & 2.8 & 0.7 & 8.9 & 5.7 & 9.5 & 1.4 & 4.9 & 16.1 & 7.2 \\
Log CR & -1.40 & -1.62 & -1.32 & -1.86 & -1.81 & -1.86 & -1.62 & -1.83 & -1.91 & -1.86 \\
\hline
\end{tabular}
\caption{Performance comparison between our proposed framework and the baseline models across 10 iterations (training objective function and log compression rate were rounded to two decimal places; train and test approximation errors were scaled to $ \times 10^{12}$). Train and test approximation errors above 1000 (after scaling) are represented by `---'. The best objective function value is designated in bold. The second best objective function value is underlined. The compression rate (CR) is the same in training and testing.}
\label{tab:results}
\end{table}

\paragraph{Interpretability Results.} To assess the interpretability of our framework, we analyse a snippet of the responses from the LLM when deciding the rank of the edge connecting mode 1 and mode 2 across the first four iterations:
\begin{itemize}
    \item \textbf{Iteration 1 Output}: \textit{`Financial instruments such as equity indexes, commodities, and currency swaps have distinct natures and market behaviors. However, within each category, the assets themselves are likely to have a lot of inter-dependencies. For instance, equity indexes are often correlated across different markets due to global economic factors. \newline 
    - Intrinsic Interaction: High \newline 
    - Suggested Rank: 5'} \newline \newline 
    \textbf{Explanation:} The LLM leverages domain knowledge to evaluate the financial instruments and provides clear explanations for the selected rank. Furthermore, the rank of 5 was chosen as it approaches the maximum allowed rank for this particular edge, which is 6, demonstrating that the LLM selects a high rank while adhering to the imposed constraints.
    \item \textbf{Iteration 2 Output}: \textit{`The interaction between the types of financial instruments and the specific assets within them is high, as previously reasoned. However, to experiment with decreasing the rank to further lower the log compression rate while keeping approximation error in check, I will reduce the rank slightly.} \newline   
    \textit{- Suggested Change: Decrease rank from 5 to 4'} \newline   \newline   
    \textbf{Explanation:} The LLM acknowledges the previous reasoning regarding why the interaction should be high, while providing a clear rationale for reducing the rank to achieve the main objective.
    \item \textbf{Iteration 3 Output}: \textit{`The interaction between types of financial instruments and specific assets within them was previously set at 4. Given that this is a high interaction, increasing the rank slightly might reduce the approximation error.} \newline 
    \textit{- Suggested Change: Increase rank from 4 to 5'}
    \newline \newline 
    \textbf{Explanation:} The LLM effectively fetches the previous used rank while demonstrating an understanding of how rank adjustments impact different components of the loss function. 
    \item \textbf{Iteration 4 Output}: \textit{`The interaction is high, and previously the rank was set at 4 which yielded better results than increasing it to 5. Let's keep it at 4.} \newline 
   \textit{ - Suggested Change:Keep rank at 4'} 
   \newline \newline 
   \textbf{Explanation:} The LLM demonstrates both reasoning and memory capabilities by recalling that a rank of 4 yielded better results than the rank of 5, and by applying its domain knowledge that a high interaction warrants a rank of 4. It effectively compares outcomes from previous iterations and understands that reducing the rank below 4 would indicate moderate interaction, aligning its decision with the overall task objective of selecting the optimal rank.
\end{itemize}
 For the following iterations, the rank was kept constant at 4, as the LLM reasons that this value best captures the interactions between the two modes while minimising the objective function. Notably, this value led to the lowest objective function at iteration 9. A similar sequence of responses can be observed for all ranks. \par 
 As a result, we find that the LLM consistently provides interpretable, task-specific, and domain-informed answers throughout the rank selection process.
 
\section{Conclusion and Future Work}
We have presented an LLM-guided tensor network rank search framework which improves the interpretability in the rank selection process. By carefully designing the prompts to help the LLM focus on the objectives and enhance its reasoning over multiple iterations, our approach has made tensor network decompositions more accessible. Experimental results on financial data tensors have shown that our framework not only enhances interpretability but also helps non-experts without the domain-specific knowledge to effectively utilise tensor network models, whilst also showcasing strong generalisation properties. This work contributes to making advanced tensor techniques more widely utilised and propels future research at the intersection of large language models and higher-order data analysis. \par 

\paragraph{Future work.} Although our proposed framework has demonstrated enhanced interpretability, we aim to further improve both the accuracy and interpretability of our LLM-guided rank selection framework. In future research, we plan to explore additional prompting techniques, such as Tree-of-Thought \cite{tree_thought}, which has proven to be powerful for complex reasoning tasks requiring multi-step thinking. Furthermore, we intend to also compare our approach with sampling-based methods, such as TnALE \cite{tnale}. Additionally, we aim to mathematically verify the reasoning of the LLMs by developing novel visualisation methods for the core tensor entries of the obtained FCTN decompositions.

\paragraph{Limitations.} The LLM-guided process relies on the reasoning consistency of the LLM, i.e., how often the LLM hallucinates, and how faithful the LLM is \cite{lanham2023measuring}. In practice, we observed that although the LLM reasons why it should not use the same set of tensor network ranks twice, it would sometimes hallucinate and output previously used tensor network rank combinations. Using LLMs specifically trained for reasoning and experimenting with techniques such as scaling up test-time compute \cite{snell2024scaling} promises to improve the proposed framework.

\bibliographystyle{unsrt}
\bibliography{references.bib}

\newpage
\appendix

\section{Mathematical definition of the FCTN decomposition}
\label{sec:fctn}
Fully Connected Tensor Network (FCTN) decomposition was proposed to allow for the correlation characterisations between any two modes \cite{fctn}. This was achieved by fully connecting every decomposed tensor. The FCTN decomposition also has the property of transpositional invariance. For an order-$N$ tensor $\mathcal{X} \in \mathbb{R}^{I_1 \times I_2 \times \cdots \times I_N}$, its Fully Connected Tensor Network decomposition can be written as \cite{fctn}
\begin{equation}
\begin{aligned}
\mathcal{X}(i_1, i_2, \dots, i_N) = &
\sum_{r_{1,2}=1}^{R_{1,2}} \sum_{r_{1,3}=1}^{R_{1,3}} \dots \sum_{r_{1,N}=1}^{R_{1,N}} \dots \sum_{r_{2,3}=1}^{R_{2,3}} \sum_{r_{2,N}=1}^{R_{2,N}} \dots \sum_{r_{N-1,N}=1}^{R_{N-1,N}} \\
& \mathcal{G}_1(i_1, r_{1,2}, r_{1,3}, \dots, r_{1,N}) \\
& \mathcal{G}_2(r_{1,2}, i_2, r_{2,3}, \dots, r_{2,N}) \cdots \\
& \mathcal{G}_k(r_{1,k}, r_{2,k}, \dots, r_{k-1,k}, i_k, r_{k,k+1}, \dots, r_{k,N}) \cdots \\
& \mathcal{G}_N(r_{1,N}, r_{2,N}, \dots, r_{N-1,N}, i_N)
\end{aligned}
\end{equation}

where $\mathcal{G}_k \in \mathbb{R}^{R_{1,k} \times R_{2,k} \times \cdots \times R_{k-1,k} \times I_k \times R_{k,k+1} \times \cdots \times R_{k,N}}$ for $k=1,2,\ldots,N$ are the decomposed order-$N$ FCTN core tensors and $\{R_{k_1, k_2}\}_{k_1 = 1, k_2 = 1}^{N, N} \ \text{where} \ R_{i,j} = R_{j,i}$ are the FCTN ranks. If we set any rank to 1, it essentially drops that connection. However, despite the FCTN decomposition enabling a stronger inter-mode correlation characterisation ability, it has a rank count which scales exponentially against the number of modes, $N$. This renders the problem of finding the optimal ranks difficult.

\section{Detailed training and testing approximation errors achieved by the baselines and our proposed framework} \label{full_errors}

We show the exact training and testing approximation errors (even those over 1000) achieved by the baseline models and our proposed framework in Table \ref{tab:results_full1} and Table \ref{tab:results_full2}. Note that the LLM-suggested rank values yielded test approximation errors consistently lower than the train approximation errors after iteration 2, whereas it is not observed in the baseline models, showing the generalisation ability of our proposed framework.

\begin{table}[h]
\centering
\renewcommand{\arraystretch}{1.2}
\setlength{\tabcolsep}{3pt}
\begin{tabular}{|>{\centering\arraybackslash}p{2.3cm}|>{\centering\arraybackslash}p{2.1cm}|>{\centering\arraybackslash}p{2.1cm}|>{\centering\arraybackslash}p{2.1cm}|>{\centering\arraybackslash}p{2.1cm}|>{\centering\arraybackslash}p{2.1cm}|}

\hline
\textbf{Iteration} & 1 & 2 & 3 & 4 & 5 \\
\hline
\multicolumn{6}{|c|}{\textbf{Random Search}} \\
\hline
Train Obj. Func. & $25.98$ & $175.44$ &  $-0.95$ & $-1.82$ & $-0.86$ \\
Train Error & $2.85$ $\times 10^{10}$& $1.78 \times 10^{11}$ & $9.81\times 10^{8}$ & $8646.1$ & $1.10 \times 10^{9}$\\
Test Obj. Func. & $29.7$  & $217.2$ & $-0.88$ & $-1.82$ &  $-0.71$ \\
Test Error & $3.22 \times 10^{10}$ & $2.20 \times 10^{11}$ & $1.05 \times 10^{9}$ & $33430.5$ & $1.25 \times 10^{9}$\\
Log CR & $-2.51$ & $-2.79$ & $-1.93$ & $-1.82$ & $-1.96$ \\
\hline
\multicolumn{6}{|c|}{\textbf{Bayesian Optimisation}} \\
\hline
Train Obj. Func. & $-1.56$ & $6.45$  & $1.00$  & $3.57$ & $-1.37$  \\
Train Error & $3.41 \times 10^{8}$ & $8.65 \times 10^{9}$ & $3.09 \times 10^{9}$ & $5.78 \times 10^{9}$ & 0.7  \\
Test Obj. Func. & $-1.53$ & $7.16$ & $1.29$ & $3.82$ & $-1.37$ \\
Test Error & $3.67 \times 10^{8}$ & $9.36 \times 10^{9}$ & $3.38 \times 10^{9}$ & $6.04 \times 10^{9}$ & $1.0$  \\
Log CR & $-1.90$ & $-2.20$ & $-2.09$ & $-2.22$ & $-1.37$ \\
\hline
\multicolumn{6}{|c|}{\textbf{Our Proposed Framework}} \\
\hline
Train Obj. Func. & $-1.40$ & $-1.62$ & $-1.32$ & $\underline{-1.86}$ & $-1.81$ \\
Train Error & $0.4$ & $1.8$ & $0.8$ & $11.9$ & $9$ \\
Test Obj. Func. & $-1.40$ & $-1.62$ & $-1.32$ & $\underline{-1.86}$ & $-1.81$ \\ 
Test Error & $0.5$ & $2.8$ & $0.7$ & $8.9$ & $5.7$ \\
Log CR & $-1.40$ & $-1.62$ & $-1.32$ & $-1.86$ & $-1.81$ \\
\hline
\end{tabular}
\caption{Performance comparison between our proposed framework and the baseline models across the first 5 iterations (training objective function and log compression rate were rounded to two decimal places; train and test approximation errors were scaled to $ \times 10^{12}$). The compression rate (CR) is the same in training and testing.}
\label{tab:results_full1}
\end{table}

\begin{table}[h]
\centering
\renewcommand{\arraystretch}{1.2}
\setlength{\tabcolsep}{3pt}
\begin{tabular}{|>{\centering\arraybackslash}p{2.3cm}|>{\centering\arraybackslash}p{2.1cm}|>{\centering\arraybackslash}p{2.1cm}|>{\centering\arraybackslash}p{2.1cm}|>{\centering\arraybackslash}p{2.1cm}|>{\centering\arraybackslash}p{2.1cm}|}

\hline
\textbf{Iteration} & 6 & 7 & 8 & 9 & 10 \\
\hline
\multicolumn{6}{|c|}{\textbf{Random Search}} \\
\hline
Train Obj. Func. & $-1.47$ & $-1.48$ & $19.04$ & $0.23$ & $243.06$ \\
Train Error & $0.2$ & $3.63 \times 10^8$  & $2.15 \times 10^{10}$ & $2.25 \times 10^9$ & $2.46 \times 10^{11}$\\
Test Obj. Func. & $-1.47$ & $-1.34$  & $22.09$ & $0.64$ & $340.11$ \\
Test Error & $0.3$ & $4.99 \times 10^8$ & $2.46 \times 10^{10}$ & $2.66 \times 10^9$  & $3.93 \times 10^{11}$ \\
Log CR & $-1.47$ & $-1.84$ & $-2.51$ & $-2.02$ & $-2.89$  \\
\hline
\multicolumn{6}{|c|}{\textbf{Bayesian Optimisation}} \\
\hline
Train Obj. Func. & $-1.84$  & $-1.78$ & $2.15$ & $-1.75$ & $-1.39$  \\
Train Error & $1.9$ & $28.8$ & $4.24 \times 10^{9}$  & $1.47 \times 10^{3}$ & $0.1$ \\
Test Obj. Func. & $-1.84$ & $-1.78$  & $2.17$ & $-1.75$ & $-1.39$  \\
Test Error & $1.7$ & $15.6$ & $4.25 \times 10^{9}$ & $2.35 \times 10^{3}$ & $0.1$ \\
Log CR & $-1.84$ & $-1.78$ & $-2.09$  & $-1.75$  & $-1.39$\\
\hline
\multicolumn{6}{|c|}{\textbf{Our Proposed Framework}} \\
\hline
Train Obj. Func. & $\underline{-1.86}$ & $-1.62$ & $-1.83$ & $\textbf{-1.91}$ & $\underline{-1.86}$ \\
Train Error  & $10.7$ & $1.8$ & $7.8$ & $22.9$ & $11.2$ \\
Test Obj. Func. & $\underline{-1.86}$ & $-1.62$ & $-1.83$ & $\textbf{-1.91}$ & $\underline{-1.86}$ \\
Test Error & $9.5$ & $1.4$ & $4.9$ & $16.1$ & $7.2$  \\
Log CR & $-1.86$ & $-1.62$ & $-1.83$ & $-1.91$ & $-1.86$ \\
\hline
\end{tabular}
\caption{Performance comparison between our proposed framework and the baseline models across the last 5 iterations (training objective function and log compression rate were rounded to two decimal places; train and test approximation errors were scaled to $ \times 10^{12}$). The compression rate (CR) is the same in training and testing.}
\label{tab:results_full2}
\end{table}

\end{document}